\documentclass[preprint,12pt, a4paper]{elsarticle}

\usepackage{amssymb}
\usepackage{tabularx}
\usepackage{float}
\usepackage{color}
\usepackage{soul}
\usepackage{todonotes}
\usepackage[symbol]{footmisc}
\usepackage{rotating, graphicx}
\usepackage{lscape}
\usepackage{lscape}
\usepackage{verbatim}

\usepackage[nolist,nohyperlinks]{acronym}
\begin{acronym}
\acro{YOLO-S}{YOLO-Software}
\acro{AI}{Artificial Intelligence}
\acro{HRI}{Human-Robot Interaction}
\acro{ML}{Machine Learning}
\acro{WoZ}{Wizard-of-Oz}
\acro{KNN}{K nearest neighbour}
\acro{API}{Application Programming Interface}
\acro{GPIO}{General-purpose input/output}
\acro{STEAM}{Science, Technology, Engineering, Art, and Mathematics}
\end{acronym}
\usepackage{listings}
\usepackage{hyperref}
\usepackage{enumitem}
\usepackage{amsmath}
\usepackage{tcolorbox}

\journal{SoftwareX}

\begin{document}
\begin{frontmatter}

\title{Software architecture for YOLO,\\a creativity-stimulating robot}

\author[label1,label2,label3]{Patr\'{i}cia Alves-Oliveira\footnote[1]{Corresponding author:\\
\textit{E-mail address:} \url{patricia\_alves\_oliveira@iscte.pt} (P. Alves-Oliveira)\\
\textit{Webpage:} \url{www.patricialvesoliveira.com}}}
\author[label3,label4]{Samuel Gomes}
\author[label5]{Ankita Chandak}
\author[label1]{Patr\'{i}cia Arriaga}
\author[label2]{Guy Hoffman}
\author[label3,label4]{Ana Paiva}

\address[label1]{ISCTE-Instituto Universit\'{a}rio de Lisboa, CIS-IUL, Lisbon, Portugal}
\address[label2]{Sibley School of Mechanical and Aerospace Engineering, Cornell University, Ithaca, NY, USA.}
\address[label3]{INESC-ID, Lisbon, Portugal.}
\address[label4]{Instituto Superior T\'{e}cnico, Universidade de Lisboa, Lisbon, Portugal}
\address[label5]{Computer Science Engineering, Cornell University, Ithaca, NY, USA.}

\begin{abstract}
YOLO is a social robot designed and developed to stimulate creativity in children through storytelling activities. Children use it as a character in their stories. This article details the artificial intelligence software developed for YOLO. The implemented software schedules through several Creativity Behaviors to find the ones that stimulate creativity more effectively. YOLO can choose between convergent and divergent thinking techniques, two important processes of creative thought. These techniques were developed based on the psychological theories of creativity development and on research from creativity experts who work with children. Additionally, this software allows the creation of Social Behaviors that enable the robot to behave as a believable character. On top of our framework, we built $3$ main social behavior parameters: Exuberant, Aloof, and Harmonious. These behaviors are meant to ease immersive play and the process of character creation. The $3$ social behaviors were based on psychological theories of personality and developed using children's input during co-design studies. Overall, this work presents an attempt to design, develop, and deploy social robots that nurture intrinsic human abilities, such as the ability to be creative.
\end{abstract}

\acresetall


\end{frontmatter}

\section{Introduction}
\begin{figure*}
    \centering
    \includegraphics[width=0.4\textwidth]{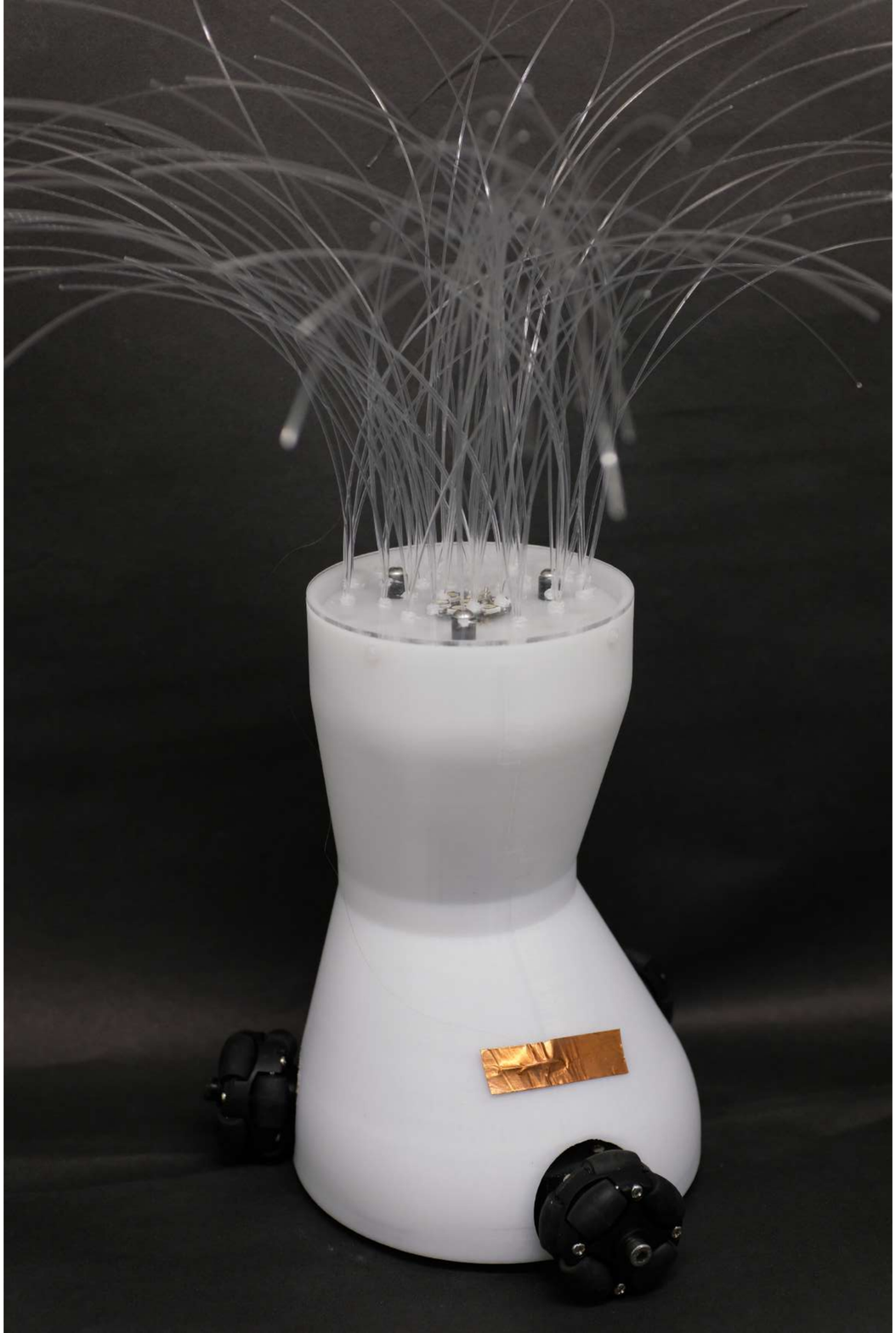}
    \caption{YOLO, a social robot that can boost creativity in children.}
    \label{fig:yolo-robot}
\end{figure*}
Creativity is one of the most sought-after skills, with recognized benefits in education, mental health, and professional success \cite{robinson2011out, collard2014nurturing}. It is associated with states of joy, play, efficiency, and pleasure \cite{baer2017content, kaufman2018creativity}. When stimulated in childhood, it promotes overall development with specific benefits in learning and adaptation. \cite{gardner2008art}. Creativity has been stimulated through the use of different intervention programs with promising effectiveness levels, demonstrating the potential to develop creativity with proper training \cite{scott2004types, ma2009effect}. However, most of these interventions developed for children lack elements of joy and fun. Therefore, they are considered similar to test-like exercises that can hinder their creative expression. The fast pace of technology development enabled the design of new tools for exploring the context of creativity stimulation \cite{shneiderman2009creativity} (e.g., the CUBUS virtual environment for storytelling with emotionally evocative characters \cite{pires2017cubus}).

In our work, we aim to expand the range of technologies used for creativity stimulation by incorporating social robots to catalyze this ability.
With this in mind, we designed and developed YOLO (Your Own Living Object), an original social robot to be used as a toy during children's play times (see Figure \ref{fig:yolo-robot}). YOLO belongs to a new generation of technological toys meant to stimulate creative abilities in children. This robot is envisioned as a character children use during storytelling. By having a small-size and a light-weight design, YOLO can be manipulated by children as if it was a traditional toy while children create stories with it (similarly to what they do with dolls or car toys). The added-value of this robot is that it can increase the creative thought process of children during story creation by providing novel ideas for storylines. Because YOLO is a non-anthropomorphic robot, it interacts with children using alternative but effective interactive modalities. These comprise variable motion and illumination profiles. In this paper, we detail the artificial intelligence software of YOLO. The software is composed of Creativity and Social Behaviors whose design was grounded on creativity research \cite{smith1998idea}, the Big Five personality model \cite{john1999big}, and co-design sessions with children \cite{alves2017yolo}.
We release the robot's software \footnote{Download YOLO Software: \url{https://github.com/patricialvesoliveira/YOLO-Software}} along with its installation guide\footnote{Link to YOLO Software installation guide: \url{https://github.com/patricialvesoliveira/YOLO-Software/wiki}} in open access. This software should be exclusively installed on YOLO hardware \cite{alves2019guide}).

\section{Related work}
\begin{figure*}
    \centering
    \includegraphics[width=0.8\textwidth]{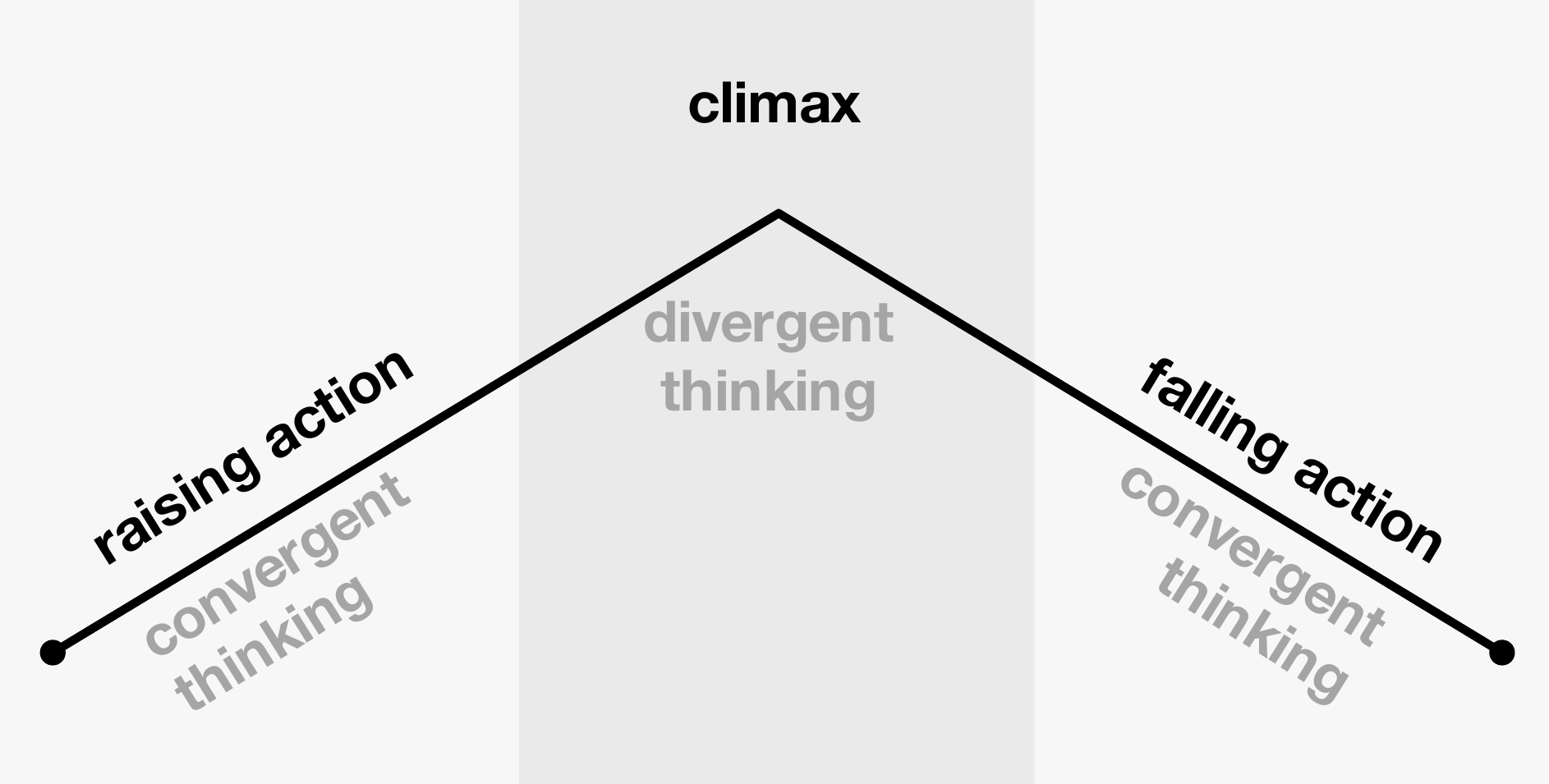}
    \caption{Storytelling arcs associated with creativity techniques: convergent thinking is stimulated during rising and falling action phases by using the mirror technique; divergent thinking is stimulated during climax by using the contrast technique.}
    \label{fig:arc}
\end{figure*}
\subsection{Background on creativity research}
\label{sec:introduction}
As societies develop into creativity-based economies, innovative and creative problem solving and the ability to collaborate are becoming must-have skills \cite{florida2005cities, robinson2011out}. However, around the age of $7$--$9$ years old there is a decline in children's creativity abilities known as the ``creative crisis'' \cite{kim2011creativity}. Research has shown that everyone has the potential to be creative and that creativity can be nurtured if stimulated \cite{runco2004everyone, sawyer2003creativity}. In our work, we aim to contribute to the increase in children's creativity by using a social robot. It therefore becomes imperative to stimulate this ability at a young age. In particular, emphasizing the school environment where children spend most of their time. While some schools already feature activities (e.g., storytelling) to support the promotion of children's creative thinking \cite{di2010collective}, current activities can be challenging to integrate into traditional classroom formats as they need preparation time and are not formally included in the school curriculum \cite{chan2014personal}. Technologies --- such as social robots --- appear as a more effective tool to apply in these contexts.


\subsection{Robots for creativity}
\label{sec:related-work}
Robots have been programmed with a deep variety of socially intelligent behaviors and affective states; thus, permitting robots to be perceived as social actors \cite{breazeal2004designing, reeves1996people}. Additionally, due to their physical and interactive nature, they become a technology that can uniquely impact creativity stimulation. Ali, Moroso, and Breazel \cite{ali2019can} demonstrated that a robot displaying creative behaviors positively influenced the creativity of children. The authors found that children who interacted with a creative robot generated more ideas, explored more themes, and were more original, than children who interacted with a non-creative robot \cite{ali2019can}. Additionally, Gordon et al. \cite{gordon2015can} demonstrated that children become more curious, an important creativity trait, when interacting with a curious robot. The authors found that these children posed more questions and become avid explores, compared to children who interacted with a non-curious robot \cite{gordon2015can}.


\section{Software description}
\label{sec:software-description}
In our work, we developed software that gives life to the social robot YOLO, whose interaction is composed of Creativity and Social Behaviors.

\begin{table}[]
\caption{Software functionalities considering the sensors used, the input collected, the actuators in place, and the output provided.}
\label{tab:software-functionalities}
\begin{tabular}{p{1.5cm}p{4.5cm}p{1.5cm}p{4.5cm}}
\hline
\textbf{Sensor} & \textbf{Input}                                         & \textbf{Actuator} & \textbf{Output}                                                                                                                                                                                                                                                                                                       \\ \hline
Touch sensor    & Ability to recognize when the robot is being touched.                  & LED lights        & The robot displays white lights while being touched, refrains from performing any behavior. When not sensing touch, the robot displays colors associated with its different social behaviors.\\
Optical sensor  & Recognition of play patterns of children while manipulating the robot. & Omni wheels       & Imitating the collected movement patterns.                                                                                  \\
Time & Stage of the storytelling that children are currently engaged in. & Omni wheels and LED lights & The robot performs a creativity technique according to the storytelling arc.\\  \hline
\end{tabular}
\end{table}

\subsection{Creativity behavior}
\label{sec:creativity-behavior}
In our specific application scenario, YOLO acts as a character that can trigger new directions in children's stories that otherwise would not emerge. During story creation, a combination of divergent (i.e., broad gathering of multiple ideas) and convergent thinking (i.e., narrowing down possibilities to create a coherent story plot) is required \cite{alrutz2015digital, brenner2016design, elbow1983teaching}.
We have chosen two techniques to stimulate creativity, named ``contrast''and ``mirror'' \cite{smith1998idea} (see Figure \ref{fig:arc})\footnote{Parameterization specifications for the creativity behaviors of YOLO can be found here: \url{https://github.com/patricialvesoliveira/YOLO-Software/wiki/CreativityProfile}}:

\begin{itemize}[noitemsep]
    \item \textbf{Contrast ---} This technique is used to stimulate divergent thinking \cite{rickards1975problem}. In the Contrast technique, YOLO provides stimuli unrelated to the storyline that children are exploring at the moment, producing an opportunity to explore new directions in the plot. This leads to heightened action and interesting plot twists in the stories of children.
    \item \textbf{Mirror ---} This technique is used to stimulate convergent thinking \cite{vangundy1988techniques}. When using the Mirror technique, YOLO provides stimuli that are connected with the storyline that children are exploring, leading to the elaboration and convergence of story ideas. This leads to then emergence of interesting details about a character, a scenario, or an action in the story.
\end{itemize}

\subsection{Storytelling arc}
\label{sec:storytelling-arc}
Successful and satisfying stories follow a storytelling arc \cite{freytag1872technik, freytag1896freytag}. According to the Theory of Dramatic Structure, each story has five acts: exposition, raising action, climax, falling action, and d\'{e}nouement \cite{freytag1872technik, freytag1896freytag}. These five acts can be modified and adapted to the dramatic structure of short stories, fables, or fairy-tales. In our software, we considered a short-story format similar to what is used in children's stories \cite{wright1995storytelling}. Therefore, we divide the narrative of a story in the following phases:

\begin{itemize}[noitemsep]
    \item \textbf{Rising action ---} Characters are introduced, a context is given to the story, and the story builds. During this stage, YOLO stimulates convergent thinking by using the mirror creativity technique (see Section \ref{sec:creativity-behavior});
    \item \textbf{Climax ---} The story reaches the point of greatest tension. During this stage, YOLO stimulates divergent thinking by applying the contrast creativity technique (see Section \ref{sec:creativity-behavior});
    \item \textbf{Falling action ---} The story shifts to an action that happens because of the climax, which means that the conflict is resolved and the story reaches its end. During this stage, YOLO stimulates convergent thinking by using the mirror creativity technique (see Section \ref{sec:creativity-behavior}).
\end{itemize}

\subsection{Social behavior}
\label{sec:social-behavior}
YOLO expresses different social profiles to exhibit social behaviors. The profiles are named \textit{Exuberant}, \textit{Aloof}, and  \textit{Harmonious}\footnote{Parameterization specifications for all the social behaviors of YOLO can be found here: \url{https://github.com/patricialvesoliveira/YOLO-Software/wiki/SocialProfile}}. These social behaviors appear as pre-sets when YOLO is turned on and can be used interchangeably, making the robot a flexible character in the children's stories. The three different social modes for YOLO are explained below:

\begin{itemize}[noitemsep]
    \item \textbf{Exuberant ---} YOLO reacts to every social interaction in an ``enthusiastic'' manner. Movements are fast and have a high amplitude. It displays vibrant colors such as purple and red with high brightness levels. As Exuberant, YOLO is proactive and seeks out social interaction. This is a vibrant, frenetic, and daring social profile;
    \item \textbf{Aloof ---} YOLO is less ``socially reactive'' and is a ``shy robot''. In this mode, the robot exhibits low amplitude, slow movements and displays cold colors such as green and blue with low brightness levels. As Aloof, YOLO is not proactive; does not seek interactions. This profile could also be described as loner, contemplative, or reclusive;
    \item \textbf{Harmonious ---} YOLO acts in a moderated fashion, presenting behaviors that are in-between the extreme versions of Exuberant and Aloof. As Harmonious, YOLO exhibits medium speed, movements with medium amplitude, and displays warm colors such as yellow and orange at medium brightness levels. This is a balanced and moderate profile.
\end{itemize}

\section{Software functionalities}
\label{sec:software-functionalities}

A primary function of this software is to serve as an \ac{API} that enables any user the opportunity to design personalized behaviors for YOLO, consequently providing the possibility to generate new behaviors and interaction modes\footnote{Guide for YOLO's \ac{API}: \url{https://github.com/patricialvesoliveira/YOLO-Software/wiki/API-Documentation}}. The robot can receive information from the environment (input) and express different interactive behaviors towards (output). Table \ref{tab:software-functionalities} lists pre-sets that were developed for YOLO to act as a social robot that can stimulate creativity in children.

Since each aspect of the robot is controllable and parameterizable, behaviors can be tweaked, created and mixed. To demonstrate the \ac{API} functionality, we conducted testing sessions in which we asked two participants unfamiliar with YOLO software to create different behaviors for the robot. One of the participants had a background in Computer Science and the other in Psychology. The participants were instructed to choose beloved characters from animation movies and to create a behavior for the robot that would resemble the behavior of those characters. The examples created by the participants were Mickey, Barbie, Bugs Bunny, and Genie from Aladdin \footnote{Examples created by the participants using the \ac{API}: \url{https://github.com/patricialvesoliveira/YOLO-Software/wiki/Examples}}.

\section{Software architecture}
\label{sec:software-architecture}
\begin{figure}
\centering
\includegraphics[width=1.0\textwidth]{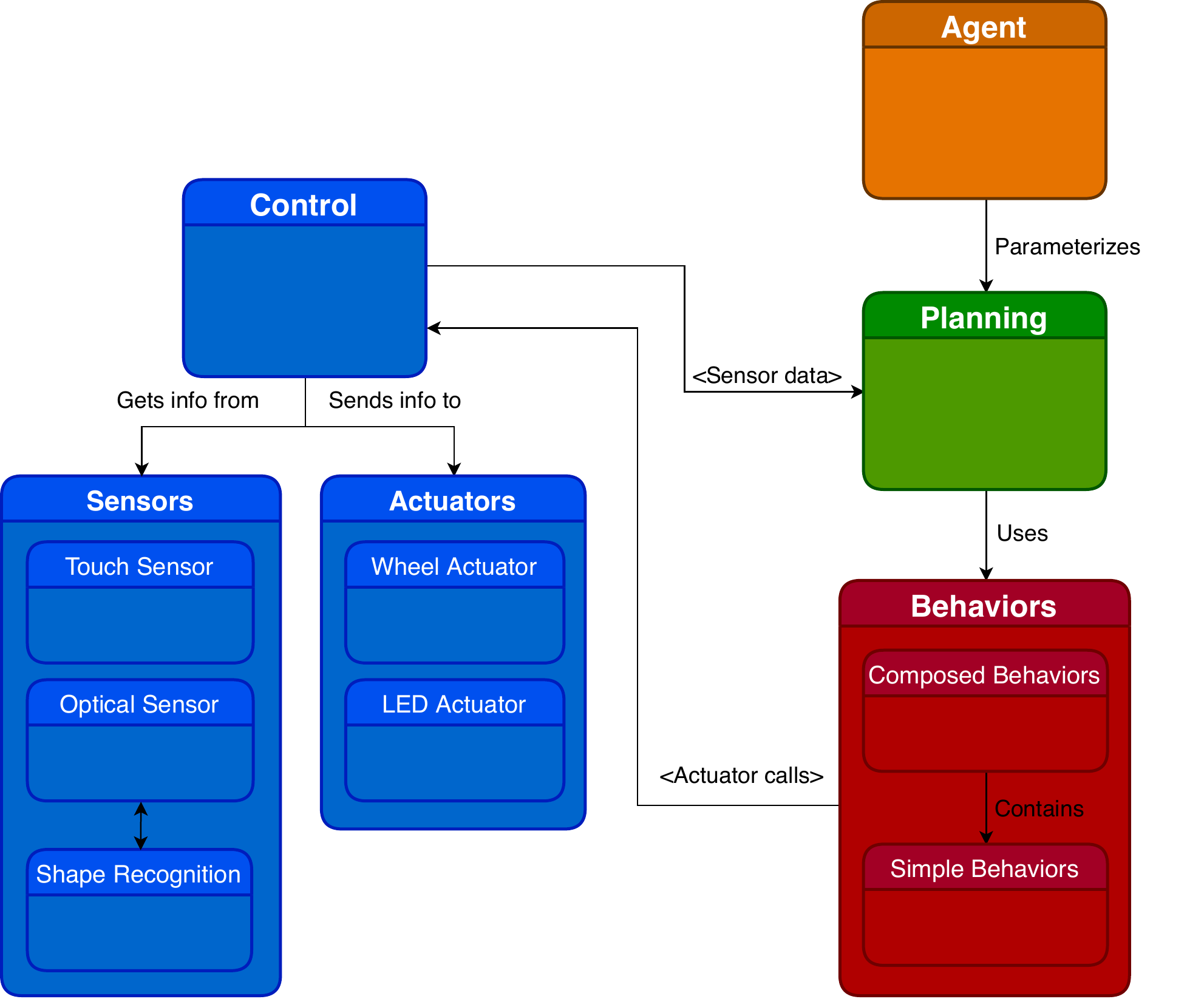}
\caption{\label{fig:YOLOModulesDiagram} Architecture of the the modules that compose YOLO software.}
\end{figure}
The architecture of our software includes several modules that manipulate data at different levels of abstraction from the low-level sensors and actuators to high-level behaviors. Figure \ref{fig:YOLOModulesDiagram} shows the scheme of these modules and how they interact. Each module is explained in the next sections.

\begin{figure}
\centering
\includegraphics[width=0.7\textwidth]{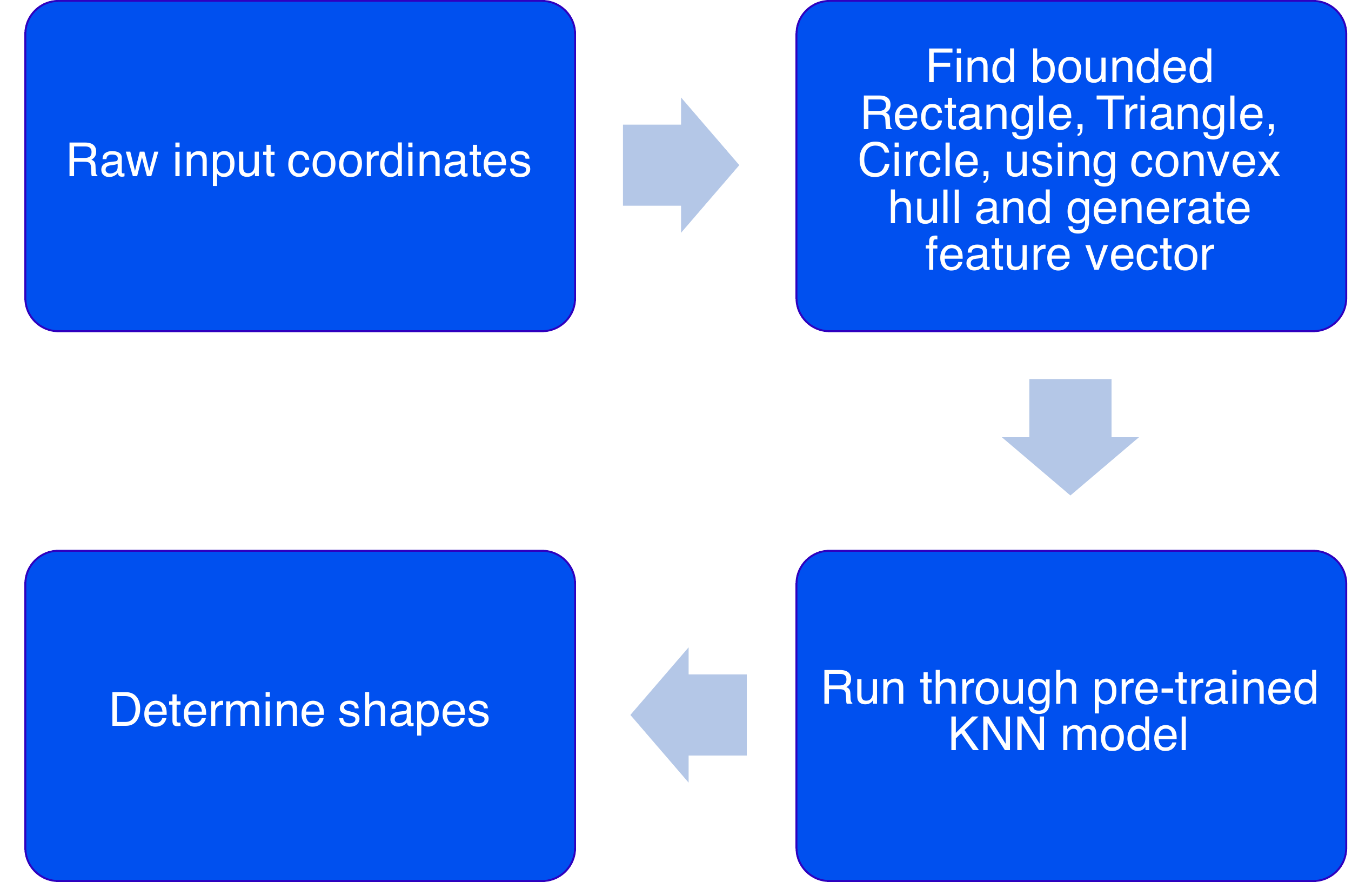}
\caption{\label{fig:YOLOShapeRecognizer} Workflow of the ML algorithm for shape recognition used by YOLO.}
\end{figure}

\subsection{Control}
\label{sec:Control-module}

This module has two main functions: first, it extracts data associated with the robots' sensors and translates into a programmable format. Second, it instructs the actuators what to do based on the software calls.

\begin{figure*}
    \centering
    \includegraphics[width=0.7\textwidth]{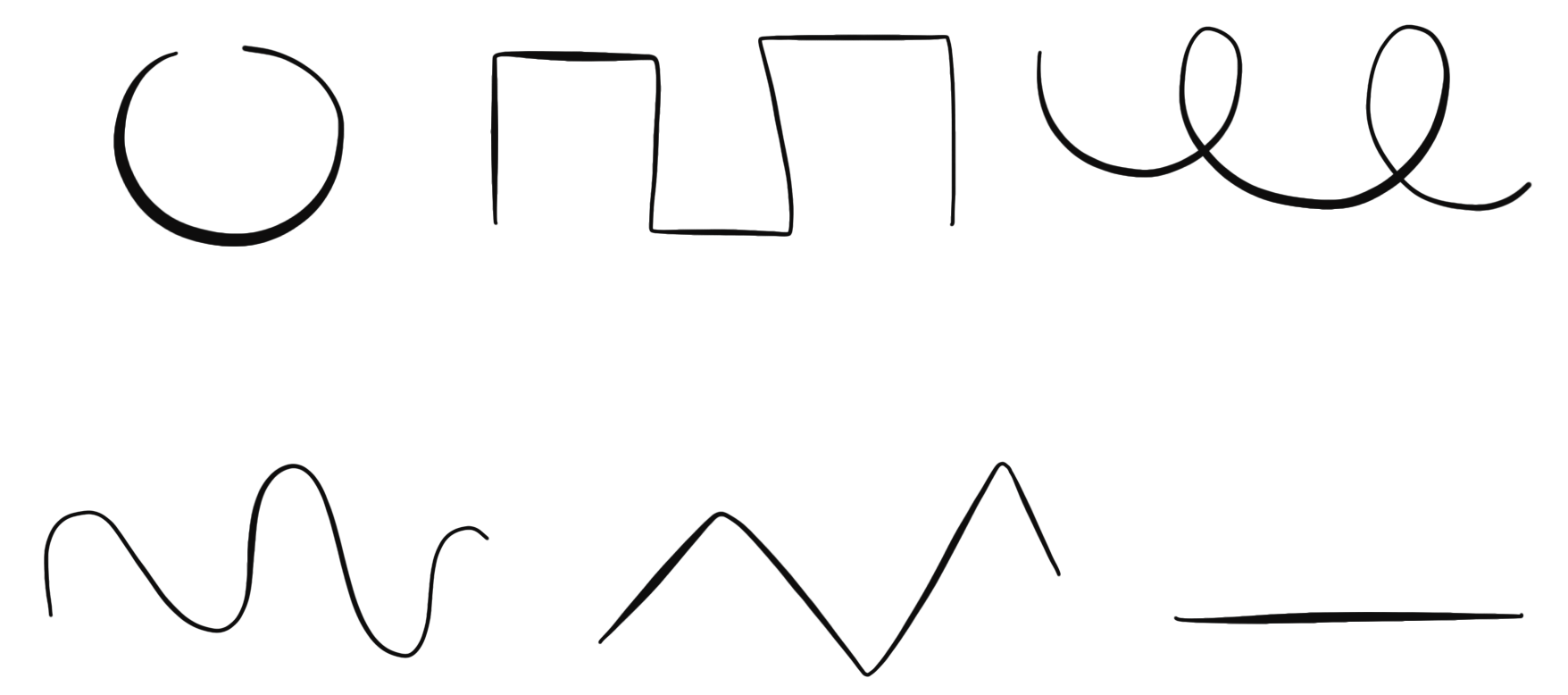}
    \caption{The lines illustrate the movement shapes performed by YOLO that are recognized by our algorithm. Illustrated shapes are: circle, rects, loops, curls, spikes, and straight line. For additional details see Section \ref{sec:Control-module}.}
    \label{fig:movement-patterns}
\end{figure*}

The \emph{touch sensor} of YOLO indicates the robot is recognizing physical contact, and the \emph{optical sensor} observes the differences in position to detect the direction of movement. The sensors record movement at each moment. The \emph{shape recognizer} dynamically identifies and characterizes each movement using \ac{ML}. The pre-trained \ac{KNN} algorithm determines a shape using the robot motion sensors which capture coordinates in $3$ seconds intervals  \cite{altman1992introduction}. Figure \ref{fig:YOLOShapeRecognizer} depicts the \ac{ML} workflow. We trained the model by collecting raw coordinates and converting these coordinates into a feature vector using the convex hull algorithm \cite{barber1996quickhull}\footnote{More details about our shape recognizer algorithm are present at this link: \url{https://github.com/patricialvesoliveira/YOLO-Software/wiki/Algorithm}}. Every time a movement is detected, \ac{KNN} is used to determine the closest matching shape from the training data. Simulations with a computer mouse showed us that with n=$3$, \ac{KNN} provided high accuracy (94\%). Therefore, we used this parameterization. The current \ac{ML} model was trained with the physical robot and can recognize with an $80$\% success rate the following shapes: circle, rectangles, loops, curls, spikes, and a straight line (see Figure \ref{fig:movement-patterns}).

YOLO actuators include the \textit{Wheel Actuator} and \textit{LED Actuator}. While \textit{Wheel Actuator} receives direction and speed values and moves the wheels' motors accordingly, \textit{LED Actuator} receives a color and brightness level and displays it in the robot's jewel LEDs.

\begin{figure}
\centering
\includegraphics[width=1.0\textwidth]{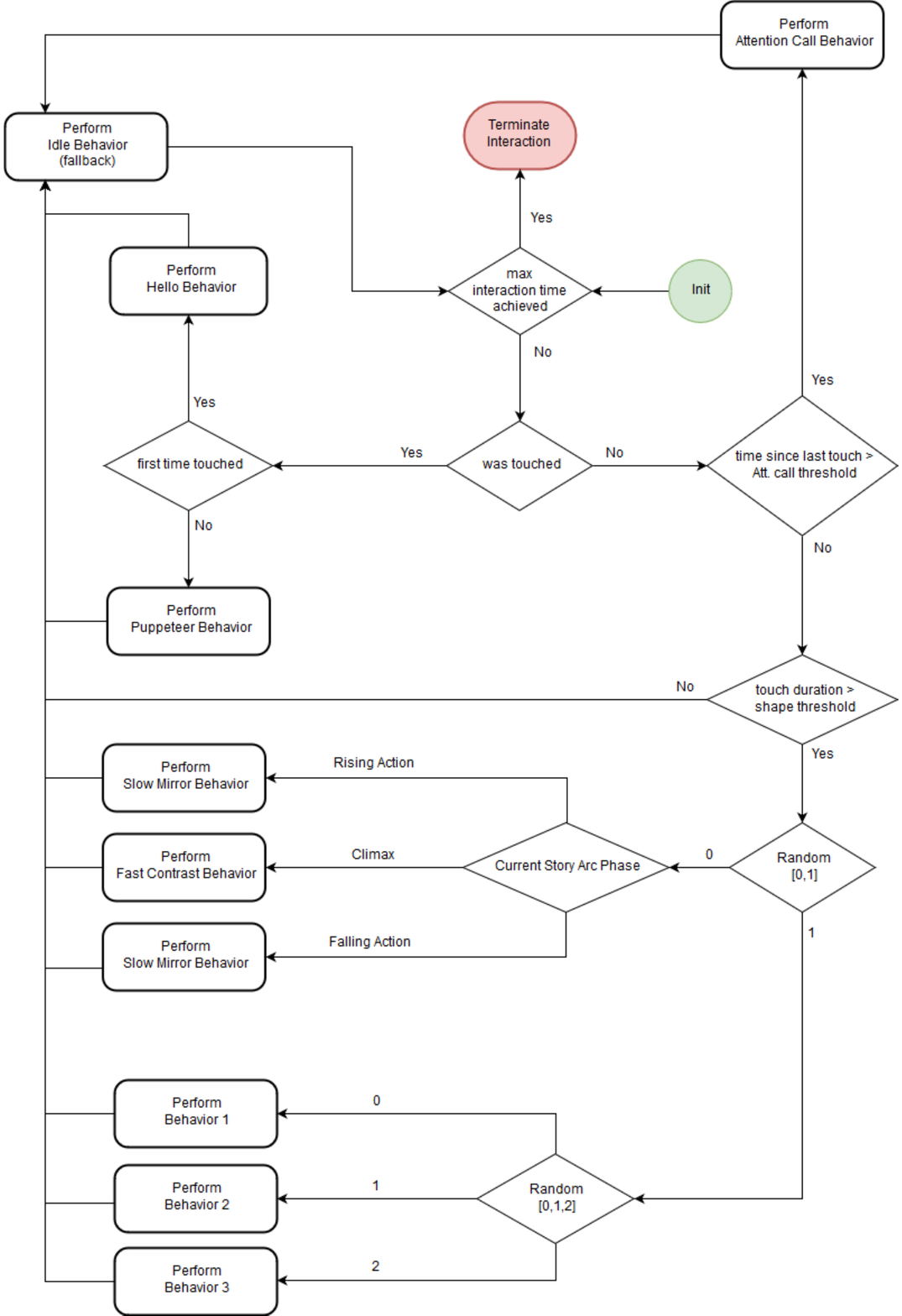}
\caption{\label{fig:PlanningFlowDiagram}State machine diagram representing the schedule of the procedures executed in the \textit{Planning} module.}
\end{figure}

\subsection{Behaviors}
The \textit{Behaviors} module coordinates the simultaneous execution of different actuators based on given parameters. The intended behavior arises from the simultaneous execution of different actuators. To simplify the development process, we divided behaviors into more concrete \textit{Simple Behaviors}, which directly use the actuator data and \textit{Composed Behaviors}, which unite several simple behaviors. \textit{Simple behaviors} directly call the \textit{Control} module. These behaviors consist of assigning different light behaviors (different colors, animations, and brightness) to different movement configurations (different movement patterns at varying speed)\footnote{Examples of simple behaviors are detailed at this link: \url{https://github.com/patricialvesoliveira/YOLO-Software/wiki/SimpleBehavior-Hierarchy}}. \textit{Composed behaviors} can be used to define the social behaviors which YOLO exhibits, such as Exuberant, Aloof, and Harmonious\footnote{Composed behaviors are further explained at this link: \url{https://github.com/patricialvesoliveira/YOLO-Software/wiki/ComposedBehavior}}.

\subsection{Planning}
The \textit{Planning} module schedules the behaviors in each moment of the interaction, executing specific ones based on the current interaction state. In order to trigger new interaction states, \textit{Planning} module uses the data extracted from the sensors which the \textit{Control} module provides. A flowchart illustrating the \textit{Planning} module's is depicted in Figure \ref{fig:PlanningFlowDiagram}.
\section{Illustrative Example}
\label{sec:illustrative-example}
\begin{figure}
\centering
\includegraphics[width=0.7\textwidth]{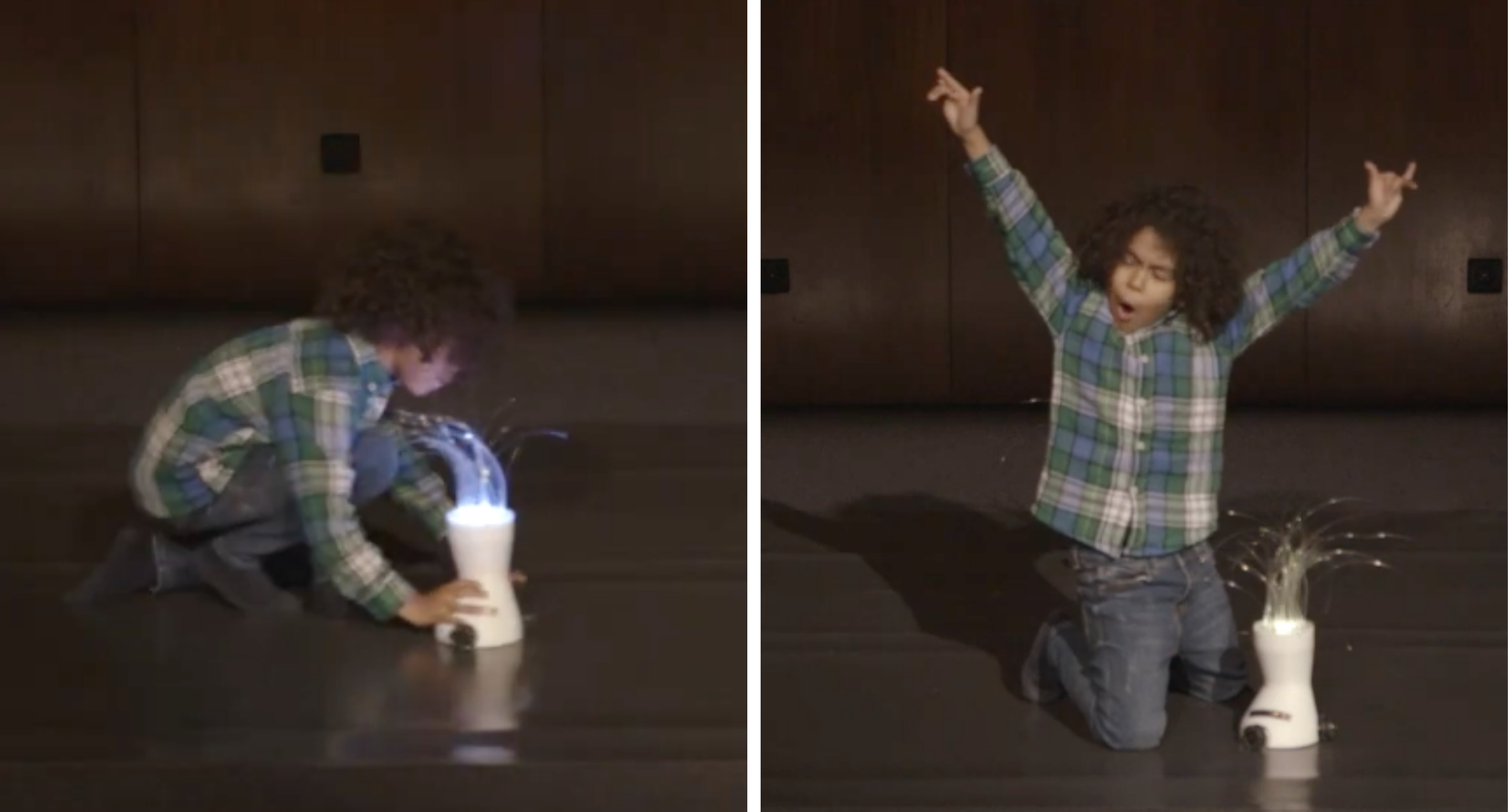}
\caption{\label{fig:use-case} Use-case example of a child using YOLO as a character for the creativity-stimulating storytelling scenario. Details about this scenario are explained in Section \ref{sec:illustrative-example}.}
\end{figure}

To validate the effectiveness of our software, we have tested it with children in a storytelling activity. The instruction provided information that they should use the robot as a character for the story they created. In the box below, we transcribed part of an interaction case during a study session between a child and YOLO (see complementary Figure \ref{fig:use-case}). In this example, it is visible how the robot makes use of its interaction profiles to stimulate convergent and divergent thinking and how this relates to the different stages of the storytelling.

\begin{tcolorbox}

The child is on the floor playing with YOLO.

\textbf{Child:} \textit{``This is a football field and YOLO is from the Benfica team, so we are going to win!''}

The child manipulates YOLO in the imaginary football field, imitating the robot running after an imaginary ball and deviating from imaginary team adversaries. Because YOLO is still in the first part of the storytelling arc, i.e., in the Raising Action stage, the robot will stimulate convergent thinking abilities. Therefore, the robot imitates the last movement that the child performed. The child looks at the robot while it is moving.

\textbf{Child:} \textit{``Yes! Go for it, C\'{a}diz, score!} (C\'{a}diz is the name of a Benfica team player that the child gave to the robot).

The child imitates scoring a goal and then grabs YOLO and celebrates.

\textbf{Child:} \textit{``Ok C\'{a}diz, but we have to continue doing well. These other guys are good too.}

The child continues manipulating YOLO through the adversaries. At this point in time, YOLO entered the next storytelling arc which is the Climax. During climax, divergent thinking is stimulated so the robot will perform a movement that is different from the last movement that  the child has performed. The child manipulates the robot straight ahead towards the soccer goal but the robot goes the opposite direction.

\textbf{Child:} \textit{``What happened? Oh no, the other guys hit you in the knee. Assistance is needed here!''}

The game continues.

\end{tcolorbox}




\section{Conclusions and Impact}
\label{sec:conclusions}
As societies develop increasingly higher levels of sophistication, social robots can play a crucial role in the development of human creativity. Related research has indicated that social robots impact the play behaviors in children, pulling them towards traditional play formats such as physical, unstructured, and unrestrained play, benefiting multiple aspects of growth \cite{pellegrini1998physical}. 

In this article, we presented the software which allows the YOLO robot to encourage creativity stimulation. This software allows potential developers to create behaviors that make the robot act according to different social behaviors. We also described several tests applying our software in real-world scenarios. These tests revealed the potential of our program, as robots using our software provoke creative narratives in stories which the children created.

The impact of this software is broad. By being an easy-to-use tool, children's stakeholders such as educators and parents, have access to a robot that is easy to prepare (e.g., for \ac{STEAM}-related activities), contrasting with other existing technological tools that can be cumbersome for non-experts to prepare \cite{chan2014personal}. Additionally, this software serves as a solid platform in academic studies, where researchers can use YOLO's \ac{API} to study child-robot interaction.
\section{Highlights}

\begin{itemize}
    \item \textbf{Creativity stimulating robot:} YOLO can stimulate children's creativity during play;
    \item \textbf{Open-access:} Access to the code and the guide to install and execute the software;
    \item \textbf{Scalability and personalization:} \ac{API} for developers to create new behaviors for YOLO;
    \item \textbf{Application:} YOLO can be used by children's stakeholders and by the research community.
\end{itemize}
\section*{Acknowledgements}
This work was supported by national funds through Funda\c{c}\~{a}o para a Ci\^{e}ncia e a Tecnologia (FCT-UID/CEC/500 21/2013) and through AMIGOS project ref. PTDC/EEISII/7174/2014. P. Alves-Oliveira acknowledges a FCT grant ref. SFRH/BD/110223/2015. We thank Andr\'{e} Pires for his contribution on the initial stage of software development.
\section*{Required Metadata}
\label{section:RequiredMetadata}

\subsection*{Current code version}
\begin{table}[!h]
\begin{tabular}{|p{6.5cm}|p{6.5cm}|}
\hline
\textbf{Code metadata description} & \textbf{Please fill in this column} \\
\hline
Current code version & v0.4 \\
\hline
Permanent link to code/repository used for this code version & \url{https://github.com/patricialvesoliveira/YOLO-Software} \\
\hline
Legal Code License   & CC Attribution 4.0 International \\
\hline
Code versioning system used & git \\
\hline
Software code languages, tools, and services used & Python \\
\hline
Compilation requirements, operating environments \& dependencies & Raspbian Stretch Lite OS\\
\hline
If available Link to developer documentation/manual & \url{https://github.com/patricialvesoliveira/YOLO-Software/wiki}\\
\hline
Support email for questions & Samuel Gomes: \url{samuel.gomes@tecnico.ulisboa.pt} and Patr\'{\i}cia Alves-Oliveira: \url{patricia_alves_oliveira@iscte-iul.pt}\\
\hline
\end{tabular}
\caption{Code metadata}
\label{} 
\end{table}

\bibliographystyle{elsarticle-num} 

\begin{thebibliography}{10}
\expandafter\ifx\csname url\endcsname\relax
  \def\url#1{\texttt{#1}}\fi
\expandafter\ifx\csname urlprefix\endcsname\relax\def\urlprefix{URL }\fi
\expandafter\ifx\csname href\endcsname\relax
  \def\href#1#2{#2} \def\path#1{#1}\fi

\bibitem{robinson2011out}
K.~Robinson, Out of our Minds: Learning to be Creative, John Wiley \& Sons,
  2011.

\bibitem{collard2014nurturing}
P.~Collard, J.~Looney, Nurturing creativity in education, European Journal of
  Education 49~(3) (2014) 348--364.
\newblock \href {http://dx.doi.org/10.1111/ejed.12090}
  {\path{doi:10.1111/ejed.12090}}.

\bibitem{baer2017content}
J.~Baer, Content matters: Why nurturing creativity is so different in different
  domains, in: Creative Contradictions in Education, vol. 1, 2017, pp.
  129--140.

\bibitem{kaufman2018creativity}
J.~Kaufman, Creativity as a stepping stone toward a brighter future, Journal of
  Intelligence 6~(2) (2018) 2--7.
\newblock \href {http://dx.doi.org/10.3390/jintelligence6020021}
  {\path{doi:10.3390/jintelligence6020021}}.

\bibitem{gardner2008art}
H.~Gardner, E.~Gardner, Art, mind, and brain: A cognitive approach to
  creativity, Basic Books, 2008.

\bibitem{scott2004types}
G.~Scott, L.~E. Leritz, M.~D. Mumford, Types of creativity training: Approaches
  and their effectiveness, The Journal of Creative Behavior 38~(3) (2004)
  149--179.
\newblock \href {http://dx.doi.org/10.1002/j.2162-6057.2004.tb01238.x}
  {\path{doi:10.1002/j.2162-6057.2004.tb01238.x}}.

\bibitem{ma2009effect}
H.-H. Ma, The effect size of variables associated with creativity: A
  meta-analysis, Creativity Research Journal 21~(1) (2009) 30--42.
\newblock \href {http://dx.doi.org/10.1080/10400410802633400}
  {\path{doi:10.1080/10400410802633400}}.

\bibitem{shneiderman2009creativity}
B.~Shneiderman, Creativity support tools: A grand challenge for hci
  researchers, in: Engineering the User Interface, Springer, London, 2009, pp.
  1--9.

\bibitem{pires2017cubus}
A.~Pires, P.~Alves-Oliveira, P.~Arriaga, C.~Martinho, Cubus: Autonomous
  embodied characters to stimulate creative idea generation in groups of
  children, in: International Conference on Intelligent Virtual Agents, Lecture
  Notes in Computer Science, vol 10498. Springer, Cham, 2017, pp. 360--373.
\newblock \href {http://dx.doi.org/10.1007/978-3-319-67401-8_46}
  {\path{doi:10.1007/978-3-319-67401-8_46}}.

\bibitem{smith1998idea}
G.~F. Smith, Idea-generation techniques: A formulary of active ingredients, The
  Journal of Creative Behavior 32~(2) (1998) 107--134.
\newblock \href {http://dx.doi.org/10.1002/j.2162-6057.1998.tb00810.x}
  {\path{doi:10.1002/j.2162-6057.1998.tb00810.x}}.

\bibitem{john1999big}
O.~P. John, S.~Srivastava, et~al., The big five trait taxonomy: History,
  measurement, and theoretical perspectives, in: Handbook of personality:
  Theory and Research, 2nd Edition, Vol.~2, The Guilford Press, New
  York/London, 1999, pp. 102--138.

\bibitem{alves2017yolo}
P.~Alves-Oliveira, P.~Arriaga, A.~Paiva, G.~Hoffman, Yolo, a robot for
  creativity: A co-design study with children, in: Proceedings of the 2017
  Conference on Interaction Design and Children, ACM, 2017, pp. 423--429.
\newblock \href {http://dx.doi.org/10.1145/3078072.3084304}
  {\path{doi:10.1145/3078072.3084304}}.

\bibitem{alves2019guide}
P.~Alves-Oliveira, P.~Arriaga, A.~Paiva, G.~Hoffman, Guide to build yolo, a
  creativity-stimulating robot for children, HardwareX 6 (2019) e00074.
\newblock \href {http://dx.doi.org/10.1016/j.ohx.2019.e00074}
  {\path{doi:10.1016/j.ohx.2019.e00074}}.

\bibitem{florida2005cities}
R.~Florida, Cities and the Creative Class, Routledge, New York, 2005.

\bibitem{kim2011creativity}
K.~H. Kim, The creativity crisis: The decrease in creative thinking scores on
  the torrance tests of creative thinking, Creativity Research Journal 23~(4)
  (2011) 285--295.
\newblock \href {http://dx.doi.org/10.1080/10400419.2011.627805}
  {\path{doi:10.1080/10400419.2011.627805}}.

\bibitem{runco2004everyone}
M.~A. Runco, Everyone has creative potential, in: Creativity: From potential to
  Realization, American Psychological Association: Washington, DC, US, 2004,
  pp. 21--30.
\newblock \href {http://dx.doi.org/10.1037/10692-002}
  {\path{doi:10.1037/10692-002}}.

\bibitem{sawyer2003creativity}
R.~K. Sawyer, M.~Csikszentmihalyi, V.~John-Steiner, S.~Moran, D.~H. Feldman,
  H.~Gardner, R.~J. Sternberg, J.~Nakamura, et~al., Creativity and Development,
  Counterpoints: Cognition, Memo, 2003.

\bibitem{di2010collective}
N.~Di~Blas, P.~Paolini, A.~Sabiescu, Collective digital storytelling at school
  as a whole-class interaction, in: Proceedings of the 9th international
  Conference on interaction Design and Children, ACM, 2010, pp. 11--19.
\newblock \href {http://dx.doi.org/10.1145/1810543.1810546}
  {\path{doi:10.1145/1810543.1810546}}.

\bibitem{chan2014personal}
S.~Chan, M.~Yuen, Personal and environmental factors affecting teachers’
  creativity-fostering practices in hong kong, Thinking Skills and Creativity
  12 (2014) 69--77.
\newblock \href {http://dx.doi.org/10.1016/j.tsc.2014.02.003}
  {\path{doi:10.1016/j.tsc.2014.02.003}}.

\bibitem{breazeal2004designing}
C.~L. Breazeal, Designing Sociable Robots, MIT press, 2004.

\bibitem{reeves1996people}
B.~Reeves, C.~Nass, How people treat computers, television, and new media like
  real people and places, CSLI Publications and Cambridge university press
  Stanford/Cambridge, MA, 1996.

\bibitem{ali2019can}
S.~Ali, T.~Moroso, C.~Breazeal, Can children learn creativity from a social
  robot?, in: Proceedings of the 2019 on Creativity and Cognition, ACM, 2019,
  pp. 359--368.

\bibitem{gordon2015can}
G.~Gordon, C.~Breazeal, S.~Engel, Can children catch curiosity from a social
  robot?, in: 2015 10th ACM/IEEE International Conference on Human-Robot
  Interaction (HRI), IEEE, 2015, pp. 91--98.
\newblock \href {http://dx.doi.org/10.1145/2696454.2696469}
  {\path{doi:10.1145/2696454.2696469}}.

\bibitem{alrutz2015digital}
M.~Alrutz, Digital storytelling and youth: Toward critically engaged praxis,
  Youth Theatre Journal 29~(1) (2015) 1--14.
\newblock \href {http://dx.doi.org/10.1080/08929092.2015.1020184}
  {\path{doi:10.1080/08929092.2015.1020184}}.

\bibitem{brenner2016design}
W.~Brenner, F.~Uebernickel, T.~Abrell, Design thinking as mindset, process, and
  toolbox, in: Design thinking for innovation, Springer, Cham, 2016, pp. 3--21.
\newblock \href {http://dx.doi.org/10.1007/978-3-319-26100-3_1}
  {\path{doi:10.1007/978-3-319-26100-3_1}}.

\bibitem{elbow1983teaching}
P.~Elbow, Teaching thinking by teaching writing, Change: The Magazine of Higher
  Learning 15~(6) (1983) 37--40.
\newblock \href {http://dx.doi.org/10.1080/00091383.1983.10570005}
  {\path{doi:10.1080/00091383.1983.10570005}}.

\bibitem{rickards1975problem}
T.~Rickards, Problem-solving through creative analysis, Wiley, 1975.
\newblock \href {http://dx.doi.org/10.1057/jors.1975.185}
  {\path{doi:10.1057/jors.1975.185}}.

\bibitem{vangundy1988techniques}
A.~B. VanGundy, Techniques of structured problem solving, Springer, 1988.

\bibitem{freytag1872technik}
G.~Freytag, Die technik des dramas, Hirzel, 1872.

\bibitem{freytag1896freytag}
G.~Freytag, Freytag's technique of the drama: an exposition of dramatic
  composition and art, Scholarly Press, 1896.

\bibitem{wright1995storytelling}
A.~Wright, Storytelling with Children, Oxford University Press, 1995.

\bibitem{altman1992introduction}
N.~S. Altman, An introduction to kernel and nearest-neighbor nonparametric
  regression, The American Statistician 46~(3) (1992) 175--185.

\bibitem{barber1996quickhull}
C.~B. Barber, D.~P. Dobkin, D.~P. Dobkin, H.~Huhdanpaa, The quickhull algorithm
  for convex hulls, ACM Transactions on Mathematical Software (TOMS) 22~(4)
  (1996) 469--483.
\newblock \href {http://dx.doi.org/10.1145/235815.235821}
  {\path{doi:10.1145/235815.235821}}.

\bibitem{pellegrini1998physical}
A.~D. Pellegrini, P.~K. Smith, Physical activity play: The nature and function
  of a neglected aspect of play, Child Development 69~(3) (1998) 577--598.
\newblock \href {http://dx.doi.org/10.1111/j.1467-8624.1998.tb06226.x}
  {\path{doi:10.1111/j.1467-8624.1998.tb06226.x}}.

\end{thebibliography}

\end{document}